\documentclass[letterpaper]{article} % DO NOT CHANGE THIS
\usepackage{aaai2026}  % DO NOT CHANGE THIS
\usepackage{times}  % DO NOT CHANGE THIS
\usepackage{helvet}  % DO NOT CHANGE THIS
\usepackage{courier}  % DO NOT CHANGE THIS
\usepackage[hyphens]{url}  % DO NOT CHANGE THIS
\usepackage{graphicx} % DO NOT CHANGE THIS
\usepackage{natbib}  % DO NOT CHANGE THIS AND DO NOT ADD ANY OPTIONS TO IT
\usepackage{caption} % DO NOT CHANGE THIS AND DO NOT ADD ANY OPTIONS TO IT

\usepackage{amssymb} % math
\usepackage{amsmath}

\usepackage{booktabs}

\usepackage{algorithm}
\usepackage{algorithmic}

\usepackage{newfloat}
\usepackage{listings}

\DeclareCaptionStyle{ruled}{labelfont=normalfont,labelsep=colon,strut=off} % DO NOT CHANGE THIS
\floatstyle{ruled}
\newfloat{listing}{tb}{lst}{}
\floatname{listing}{Listing}
\nocopyright % -- Your paper will not be published if you use this command
\title{Adaptive Spatial Goodness Encoding: Advancing and Scaling \\ Forward-Forward Learning}
\author {
    Qingchun Gong\textsuperscript{\rm 1},
    Robert Bogdan Staszewski\textsuperscript{\rm 1},
    Kai Xu\textsuperscript{\rm 2}
}
\affiliations {
    \textsuperscript{\rm 1}University College Dublin\\
    \textsuperscript{\rm 2}King's College London\\
    qingchun.gong@ucdconnect.ie,
    robert.staszewski@ucd.ie, 
    kai.xu@kcl.ac.uk
}

\usepackage{bibentry}
\begin{document}

\maketitle

\begin{abstract}
The Forward-Forward (FF) algorithm offers a promising alternative to backpropagation (BP). Despite advancements in  recent FF-based extensions, which have enhanced the original algorithm and adapted it to convolutional neural networks (CNNs), they often suffer from limited representational capacity and poor scalability to large-scale datasets, primarily due to exploding channel dimensionality. In this work, we propose adaptive spatial goodness encoding (ASGE), a new FF-based training framework tailored for CNNs. ASGE leverages feature maps to compute spatially-aware goodness representations at each layer, enabling layer-wise supervision. Crucially, this approach decouples classification complexity from channel dimensionality, thereby addressing the issue of channel explosion and achieving competitive performance compared to other BP alternatives. ASGE outperforms all other FF-based approaches across multiple benchmarks, delivering test accuracies of 99.65\% on MNIST, 93.41\% on FashionMNIST, 90.62\% on CIFAR-10, and 65.42\% on CIFAR-100. Moreover, we present the first successful application of FF-based training to ImageNet, with Top-1 and Top-5 accuracies of 51.58\% and 75.23\%. Furthermore, we propose three prediction strategies to achieve flexible trade-offs among accuracy, parameters and memory usage, enabling deployment under diverse resource constraints. 

% By entirely eliminating BP and significantly narrowing the performance gap with BP-trained models, the ASGE framework establishes a viable foundation toward scalable BP-free CNN training.

\end{abstract}

\begin{links}
    \link{Code}{https://github.com/KAI-Laboratory/ASGE}
\end{links}

\section{Introduction}

Artificial Neural Networks (ANNs) have become a cornerstone of modern machine learning, largely owing to the effectiveness of the backpropagation (BP) algorithm \cite{ru:86}.
%serving as a training algorithm.
BP has driven the success of ANNs across a diverse range of applications \cite{l:14,ka:15,l:15,d:17,as:23} by enabling efficient training through gradient-based optimization. The algorithm proceeds in two phases: a forward pass computes the network's predictions and their associated error relative to the ground truth, followed by a backward pass that propagates this error through the network. This yields gradients for each layer's parameters (weights and biases), which are then used by optimization methods such as stochastic gradient descent (SGD) \cite{b:10} to iteratively minimize the prediction error. However, BP suffers from inherent limitations. It requires symmetric weight matrices for the forward and backward passes to compute parameter gradients via the chain rule \cite{g:87}. Additionally, it necessitates the storage of intermediate activations until parameter updates occur \cite{w:17}. These constraints incur substantial computational and memory overhead, while simultaneously restricting the degree of parallelization, ultimately rendering BP inefficient and ill-suited for deployment on resource-constrained hardware. 

In response, numerous backpropagation-free (BP-free) strategies have emerged to address these limitations, including Target Propagation (TP) \cite{y:14}, Feedback Alignment (FA) \cite{l:16}, Direct Feedback Alignment (DFA) \cite{a:16}, Deep Random Target Projection (DRTP) \cite{f:21}, and Recursive Local Representation Alignment (rLRA) \cite{o:23}, alongside other biologically inspired or local learning methods \cite{h:02,m:22,g:23}. Although these approaches alleviate some of the BP limitations, they continue to face challenges in balancing the trade-offs between prediction accuracy, memory efficiency and computational cost.

\begin{figure}
	\centering
	\includegraphics[width=0.9\columnwidth]{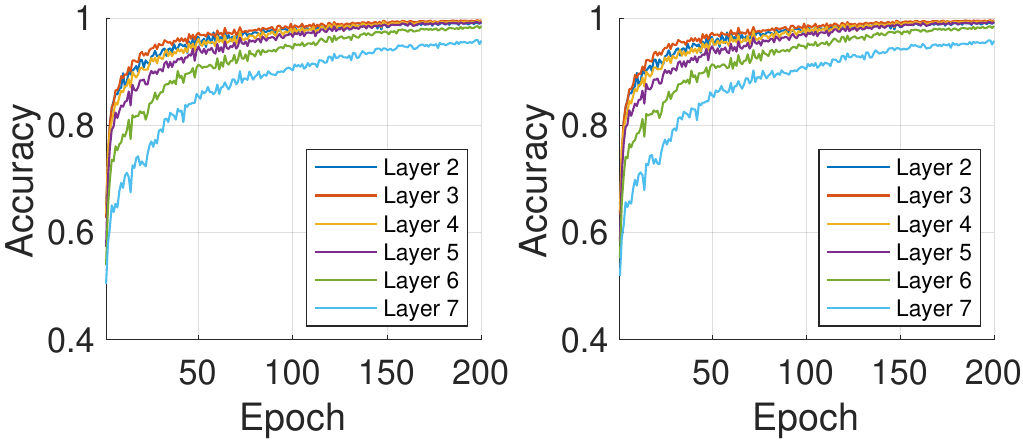} % Reduce the figure size so that it is slightly narrower than the column.
	\caption{Layer-wise training accuracy (layers 2-7) of VGG8 trained with FF on CIFAR-10. First and final layers excluded. Left: positive samples. Right: negative samples.}
	\label{fig1}
\end{figure}

A recent promising alternative is the Forward-Forward algorithm (FF) \cite{g:22}, which employs a layer-wise training strategy through dual forward passes: one with positive samples and the other with negative samples. Each layer independently updates its parameters to maximize a ``goodness" metric for positive inputs while minimizing it for negative ones. By eliminating the need for backward passes, FF significantly reduces memory overhead and facilitates pipe-lined parallelism. However, Figure~\ref{fig1} illustrates a critical limitation of FF when applied to convolutional neural networks (CNNs): prediction accuracy tends to deteriorate progressively with network depth. This indicates a weakening capacity for feature abstraction in deeper layers, thereby constraining the algorithm's effectiveness in deep architectures and on complex tasks. Recent enhancements, such as Channel-wise Competitive (CwC) learning \cite{p:24} and DeeperForward \cite{l:25}, attempt to mitigate this issue. While partially effective, they introduce an undesirable coupling between channel dimensionality and classification complexity. Specifically, they require layer-wise channels $C_l$ to scale linearly with classes $N$, i.e., $C_l \in \Theta(N)$. This scaling becomes impractical for large-scale datasets like ImageNet \cite{o:15} due to the dramatic increase in channel dimensionality.

To address both the representational degradation and scalability bottleneck in existing FF-based methods, we propose an enhanced FF-based training framework that simultaneously improves representational capacity and scalability to large-scale datasets. Our key contributions are as follows: 
\begin{itemize}
\item Adaptive Spatial Goodness Encoding (ASGE) Framework: A novel BP-free training method that decouples channel dimensionality from classification complexity, enforcing constant channel scaling ($C_l \in \Theta(1)$). This enables efficient and scalable optimization on large-category datasets.
\item ImageNet Breakthrough: Comprehensive experiments across various image classification tasks, and first successful scaling of FF-based methods to ImageNet.
\end{itemize}

\section{Background}

BP \cite{ru:86} enables end-to-end optimization by propagating errors from the output layer back through the network. However, it suffers from several inherent limitations. First, it requires weight symmetry between the forward and backward passes \cite{g:87}. Second, BP introduces the update locking problem \cite{w:17}, wherein weight updates cannot be performed until the entire forward pass is completed. These constraints result in high memory overhead due to the need to store intermediate activations and hinder layer-wise parallelism, ultimately rendering BP inefficient for deployment on resource-constrained hardware.

To overcome the limitations of backpropagation, numerous alternatives have been proposed that either avoid or approximate its core mechanism. Feedback Alignment (FA) \cite{l:16} eliminates the requirement for weight symmetry in the backward pass by using fixed, randomly initialized feedback weights. DFA \cite{a:16} extends this by propagating output layer errors to hidden layers, bypassing the backward pathway altogether. Another related method, rLRA \cite{o:23} enables weight updates without computing global gradients by recursively aligning each layer's output with a synthetic target derived from the layer above. PEPITA \cite{g:23} replaces the backward pass with a second forward pass, in which the input is modulated according to the output-layer errors from the first forward pass. Notably, these approaches still necessitate full forward passes to compute global errors or store intermediate activations prior to updating weights. To eliminate update locking, 
local learning rules---which allow layers to update independently without waiting for the entire forward pass---are desirable. DRTP \cite{f:21} achieves this by directly projecting true labels to each layer, generating layer-wise error signals that support local updates without waiting for the full forward pass. Another approach is Hebbian learning \cite{h:02}, a biologically inspired local learning rule that strengthens synaptic connections when pre-synaptic and post-synaptic neurons are activated simultaneously. Despite its biological plausibility, Hebbian learning does not support supervised learning or error correction. SoftHebb \cite{m:22,a:23} extends Hebbian learning by incorporating soft winner-take-all dynamics and differentiable surrogate gradients, enabling the training of deeper networks while preserving the characteristics of local learning. Similarly, Predictive Coding \cite{r:99,b:22} offers a local learning framework in which each layer predicts the activity of the layer below, using the resulting prediction error to locally update both prediction and representation neurons.

The FF algorithm \cite{g:22} introduces a fundamentally different training paradigm that entirely avoids the need for backpropagation. It employs a layer-wise learning mechanism in which each layer is independently supervised through two distinct forward passes---one with positive data and the other with negative data.
% During training, each layer is encouraged to learn features that distinguish whether the input corresponds to a positive or negative sample, thereby enabling the network to develop representations useful for classification. 
 Each layer computes a ``goodness" score, defined as the sum of squared activations, and is trained using a binary cross-entropy loss to classify inputs as positive if the goodness exceeds a predefined threshold, and negative otherwise. Several extensions have been proposed to enhance the performance of FF. Contrastive Forward-Forward \cite{h:25} redefines goodness as the dot-product similarity between representations of different samples and introduces layer-wise supervised contrastive learning to improve the discriminative power of the network. SymBa \cite{h:23} addresses the inherent asymmetry in binary cross-entropy loss by introducing a symmetric loss function with respect to positive and negative data, leading to improved convergence stability.  SFFA \cite{e:25} further enforces symmetry by dividing each layer's neurons into positive and negative subsets, selectively activating them based on the input's polarity. To overcome the lack of inter-layer interaction in FF, Layer Collaboration \cite{gu:23} proposes a mechanism that allows goodness information to flow across layers, improving representational coherence and overall accuracy. However, these variants have not yet been extended to CNNs, limiting their applicability to more complex classification tasks. 

\begin{figure*}[t]
\centering
\includegraphics[width=0.9\textwidth]{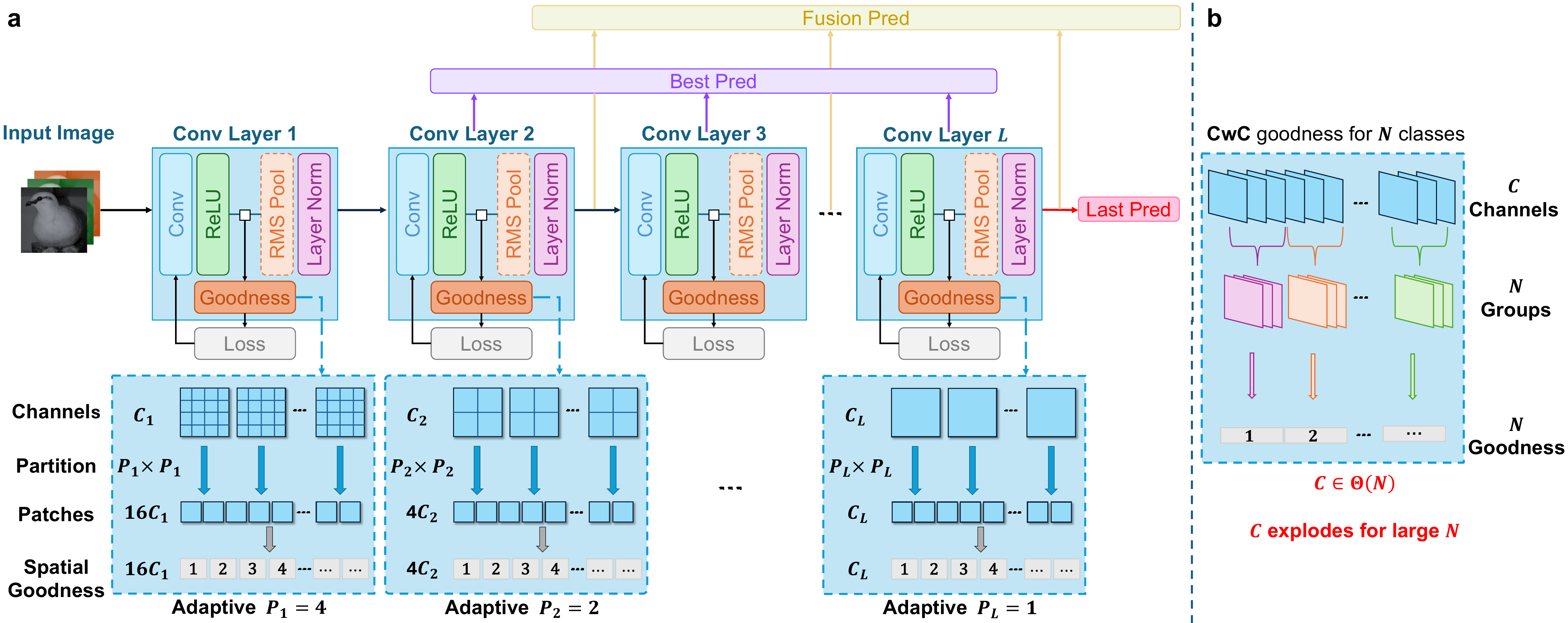} % Reduce the figure size so that it is slightly narrower than the column.
\caption{(a) Top: ASGE architecture with prediction strategies (\emph{Fusion Pred}: Cumulative Layer Fusion Prediction, \emph{Best Pred}: Best Layer Selection Prediction, \emph{Last Pred}: Last Layer Only Prediction); Bottom: Layer-wise adaptive spatial goodness. (b) CwC goodness.}
\label{fig2}
\end{figure*}

Several recent efforts have explored adapting the FF paradigm to CNNs \cite{r:24, x:25} by injecting polarity information into the global image space to generate positive and negative samples instead of using one-hot label embeddings as in the original FF. Other approaches take this further by eliminating the reliance on explicitly generated negative samples. For example, Cascaded Forward (CaFo) \cite{z:23} first trains the convolutional backbone using DFA, followed by training a set of layer-specific predictors, both stages relying solely on the original data. However, it suffers from slow convergence. A more recent state-of-the-art method is CwC learning \cite{p:24}, which partitions the output channels of each layer based on the number of target classes. It calculates a group-wise goodness score for each class-aligned channel group and then uses these scores for classification (Figure~\ref{fig2}). CwC has been shown to progressively enhance class-specific feature activation in deeper layers, partially addressing the representation degradation (Figure~\ref{fig1}). DeeperForward \cite{l:25} further extends CwC to deeper convolutional architectures and introduces a parallel training strategy to enhance efficiency. However, a significant limitation of CwC lies in its tight coupling between the number of channels and the number of classes. This design causes the channel dimensionality to scale linearly with the number of categories, severely limiting their performance on large-scale benchmarks and rendering them impractical for scaling to ImageNet. In light of these challenges, our work aims to further optimize FF-based training approaches for CNNs, with the goal of achieving improved scalability and performance on large-scale classification tasks.

\section{Method}

This section presents a backpropagation-free training framework built upon the proposed ASGE, a new mechanism that enhances representational capacity and overcomes the scalability limitations of prior FF-based methods in large-scale image classification tasks. ASGE trains each convolutional layer independently through channel-adaptive spatial goodness supervision, completely eliminating backward passes. We first introduce the core design of the ASGE mechanism, then detail the overall training framework.

\subsection{Adaptive Spatial Goodness Encoding}

\subsubsection{Spatial Goodness Extraction}

Figure~\ref{fig2} illustrates the fundamental difference in goodness vector construction between our approach and CwC \cite{p:24}. We propose a spatially grounded, class-agnostic method, where spatial goodness is defined as the energy within localized regions of each channel. This design explicitly decouples goodness computation from class count, enabling flexible supervision while ensuring scalability to large classification tasks. Formally, the feature maps of a given layer $l$ are denoted as $\mathbf{Y}_l \in \mathbb{R}^{C_l \times H_l \times W_l} $ (post-activation), where $C_l$ represents channels and $H_l \times W_l$ spatial dimensions. To standardize the quantification of localized regions, each $H_l \times W_l$ feature map is evenly partitioned into $P_l \times P_l$ patches. Accordingly, $\mathbf{Y}_l$ is reshaped into:
\begin{equation}
    \mathbf{\tilde{Y}}_l \in \mathbb{R}^{C_l \times P_l \times P_l \times H^{'}_l \times W^{'}_l }
\end{equation}
where $H^{'}_l=H_l/P_l$, $W^{'}_l=W_l/P_l$. The spatial goodness of each patch is computed as the mean squared activation (energy) within the patch region:
\begin{equation}
    g_{l,c}^{(i,j)} = \frac{1}{H^{'}_lW^{'}_l}\sum_{h=1}^{H^{'}_l} \sum_{w=1}^{W^{'}_l} \left(\mathbf{\tilde{Y}}_l^{\left(c,i,j,h,w\right)}\right)^{2}
\end{equation}
where $g_{l,c}^{(i,j)}$ denotes the spatial goodness of the $(i,j)$-th patch in channel $c$ at layer $l$. The spatial goodness values of all patches across all channels are concatenated to form the goodness vector $\mathbf{g}_{l} \in \mathbb{R}^{1 \times \left(C_l \cdot P_l^{2}\right)}$.

\subsubsection{Channel-Aware Patch Partitioning}

As discussed above, spatial goodness extraction relies on partitioning feature maps into patches. However, applying a fixed partitioning scheme uniformly across all layers overlooks the varying levels of feature abstraction captured at different depths of a CNN \cite{m:13}. To address this, we propose a channel-aware patch partitioning strategy, in which a partitioning factor $P_l$ is adaptively scaled inversely with channel count $C_l$. Specifically, shallow layers tend to capture low-level, fine-grained features and thus benefit from finer spatial partitioning to produce sufficiently expressive goodness vectors for supervision. In contrast, deeper layers encode more abstract, high-level representations, where coarser spatial partitioning is sufficient to yield meaningful goodness vectors. This behavior is consistent with CNN architectures such as VGG \cite{k:15} and ResNet \cite{ka:15}, where early layers typically have fewer channels and higher spatial resolutions, while deeper layers exhibit increased channel dimensionality and reduced spacial resolution. Consequently, it is intuitive to decrease $P_l$ as $C_l$ increases, as shown in Figure~\ref{fig2}. We therefore compute $P_l$ as:
\begin{equation}
    P_l = \min \left(\max \left(1,\left\lfloor\frac{\alpha \cdot C_L}{C_l}\right \rfloor\right), H_l, W_l \right)
\label{formula3}
\end{equation}
where $C_L$ is the channel count of the final convolutional layer, and $\alpha \ge 0$ is a hyperparameter that controls partitioning granularity. This formulation ensures that $P_l \ge 1$ , $H^{'}_l \ge 1$ and $W^{'}_l \ge 1$, while adaptively reducing the number of patches $P_l$ as channel dimensionality $C_l$ increases. A larger $\alpha$ results in finer partitioning with more patches, allowing the model to capture more detailed spatial goodness patterns. However, excessively large $\alpha$ may not yield further performance improvements. The effect of varying $\alpha$ will be further examined in the experimental section.

\subsubsection{Random Projection for Supervision}

The spatial goodness vectors obtained as described above are used for layer-wise supervision via classification. A common approach is to employ auxiliary classifiers \cite{e:19, a:19}; however, this introduces additional parameters and requires local backpropagation through the auxiliary branches, undermining the goal of achieving fully backpropagation-free training. To address this, we adopt a fixed random projection approach inspired by prior works \cite{h:06,h:16,a:25}. Specifically, each layer's goodness vector is projected into the $N$-dimensional class space via a \emph{linear} transformation defined by a weight matrix $\mathbf{W}_l \in \mathbb{R}^{\left(C_l \cdot P_l^2 \right)\times N}$ and a bias vector $\mathbf{b}_l \in \mathbb{R}^{1\times N}$, both of which are randomly initialized at the beginning of training and remain fixed throughout. The generation procedure for these weight and bias parameters is as follows:
\begin{equation}
    \mathbf{W}_l \sim \mathcal{N}\left(0,\frac{1}{N} \right),    \mathbf{b}_l \sim \mathcal{N}\left(0,\frac{1}{N} \right)
\end{equation}
The scaling factor $1/N$ for $\mathbf{W}_l$ follows the Johnson-Lindenstrauss lemma, ensuring that the $\ell_2$ norm of the projected vector remains close to that of the original goodness vector \cite{j:84}. The bias vector $\mathbf{b}_l$ is similarly scaled by $1 /N$, serving as a static offset that improves the logit distribution after projection. The classification logits $\mathbf{a}_{l} \in \mathbb{R}^{1 \times N}$ are obtained from $\mathbf{g}_{l}$ via:
\begin{equation}
    \mathbf{a}_{l} = \mathbf{g}_{l}\mathbf{W}_l+\mathbf{b}_l
\end{equation}
which is then passed through a local cross-entropy loss $\mathcal{L}_{l}\left( \mathbf{a}_{l};t^\ast\right)$, computed against the ground-truth target $t^\ast$: 
\begin{equation}
    \mathcal{L}_{l}\left( \mathbf{a}_{l};t^\ast\right) =-\mathbf{a}_{l}^{\left(t^\ast\right)} + \text{log}\sum_{n=1}^{N}{\text{exp}\left(\mathbf{a}_{l}^{\left(n\right)}\right)}
\end{equation}
The loss is subsequently used to compute gradients with respect to the convolutional parameters $\boldsymbol{\theta}_{l}$ of layer $l$:
\begin{equation}
    \nabla_{\theta_{l}}\mathcal{L}_{l} =
    \left(\frac{\partial \mathcal{L}_{l}}{\partial \mathbf{a}_{l}}\mathbf{W}_l^\top\right)
    \frac{\partial\mathbf{g}_{l}}{\partial \mathbf{Y}_{l}}\frac{\partial \mathbf{Y}_{l}}{\partial\boldsymbol{\theta}_{l}}
\end{equation}
This formulation allows the loss to directly produce gradients for the convolutional parameters, eliminating the need for local backpropagation through auxiliary trainable layers.

\subsection{Training Framework}

\subsubsection{Overview}

As illustrated in Figure~\ref{fig2}, during one forward pass, each layer is trained independently using spatial goodness-based supervision. Specifically, the convolutional output at each layer is first passed through a ReLU nonlinearity. The resulting activated feature maps are then used to compute the layer's spatial goodness and the corresponding loss, while simultaneously being forwarded to the subsequent operations---pooling (if present) and layer normalization
%\footnote{Following Hinton's official FF implementation, the layer normalization is implemented as root mean square (RMS) normalization, which normalizes activations by their RMS.}
\cite{j:16}. Critically, each layer's output is detached before propagation to downstream layers, preventing gradient flow across layers and maintaining strict layer-wise independence.

\subsubsection{Goodness-Preserving Pooling} Pooling is a standard downsampling operation in CNNs, used to reduce spatial resolution and computational complexity. However, to preserve spatial goodness information during downsampling, the pooling method must align with the energy-based formulation used to compute goodness. Standard approaches like max pooling and average pooling distort energy representation: the former tends to exaggerate energy by focusing on extreme values, while the latter may suppress it by smoothing activations. To maintain compatibility with our energy-based goodness metric, we adopt an approach inspired by Learned-Norm Pooling  \cite{c:14} which generalizes pooling through the use of $L_p$ norm. Specifically, we set $p$ = 2, resulting in a root mean square (RMS) pooling. Given a localized region $\mathcal{R} \in \mathbb{R}^{H \times W}$ of a feature map, RMS pooling is defined as:
\begin{equation}
    \text{RMSPool}\left(\mathcal{R}\right) = \sqrt{\frac{1}{HW}\sum_{h}^{H}\sum_{w}^{W}\left(\mathcal{R}^{\left(h,w\right)}\right)^{2}}
\end{equation}
RMS pooling preserves the fundamental energy relationship, which directly aligns with our mean-squared-activation-based goodness computation, thus serves as an effective goodness-preserving downsampling mechanism. 

\subsubsection{Prediction Strategies} We introduce three strategies (Figure~\ref{fig2}) for generating the final prediction during training and inference: (1) Last Layer Only prediction (\emph{Last Pred}), (2) Cumulative Layer Fusion prediction (\emph{Fusion Pred}), and (3) Best Layer Selection prediction (\emph{Best Pred}).

In the \emph{Last Pred} strategy, only the final convolutional layer's features are passed through global average pooling (GAP), flattened, and fed into a single-layer linear classifier. During training, the resulting loss is used exclusively to update the classifier's parameters, leaving earlier convolutional layers unaffected.

Given that each convolutional layer in our framework receives direct supervision from the ground-truth labels, each is expected to learn features that are independently useful for classification. To capitalize on this, the \emph{Fusion Pred} strategy aggregates information from all convolutional layers except the first---excluded due to its low prediction accuracy. Specifically, we apply GAP and flattening to the feature maps of each selected layer, then concatenate the resulting representations. This combined vector is then passed to a single-layer linear classifier. Similar to the Last Pred strategy, only the classifier's parameters are updated by the corresponding loss during training. This strategy effectively captures a broader range of features by leveraging the predictive capacity of multiple layers. However, it introduces additional classifier parameters and incurs memory overhead due to the need to store intermediate representations. This approach draws inspiration from the original FF method introduced in \cite{g:22}, but is specifically adapted to suit a convolutional architectures.

In the \emph{Best Pred} strategy, each convolutional layer produces a classification output and associated prediction accuracy during training. The final prediction is then selected from the layer that achieves the highest validation accuracy. This approach offers two key advantages: (1) it eliminates the need for a final linear classifier, and (2) it enables early exit from inference when the most accurate prediction arises from a shallower layer, reducing computational cost.

\section{Experiments}

We evaluate our approach on five benchmark datasets: MNIST \cite{l:98}, FashionMNIST (F-MNIST) \cite{h:17}, CIFAR-10,  CIFAR-100 \cite{k:09}, and ImageNet (ILSVRC 2012). To assess scalability, we include ImageNet's 1.28 million training and 50,000 validation images across 1000 classes. For MNIST and F-MNIST, the original training sets are split into 50,000 training and 10,000 validation samples, and models are trained for 100 epochs. For CIFAR-10 and CIFAR-100, we use 45,000 training and 5,000 validation samples, with training conducted over 400 epochs. For all datasets except ImageNet, we adopt the AdamW optimizer with a weight decay of 0.001, a batch size of 128 and an initial learning rate of 0.0002, which is decayed to 0.00001 using a cosine annealing scheduler. For ImageNet, we use the SGD optimizer with momentum 0.9 and weight decay 0.0001. The batch size is 512, and the learning rate is scheduled from 0.05 to 0.0005 via cosine annealing. Models are trained for 45 epochs. Training is performed on 2 NVIDIA A100 GPUs (80 GB) and 1 Intel(R) Xeon(R) Gold 6338 CPU. We report the mean accuracy and standard deviation averaged over three independent runs. In each run, the model checkpoint with the highest validation accuracy is selected for final evaluation on the test set. Additional details, including model architectures, data augmentation strategies and so on, are provided in the supplementary materials.

\subsection{Ablation Study}

%The ablation studies are to analyze the impact of key hyper-parameters and training framework components mentioned in the method section. We begin by examining how ASGE improves layer-wise prediction accuracy against FF as shown in Figure \ref{fig1}, where each positive sample is the original image and each negative sample is generated by mixing two images of different classes through a binary mask \cite{g:22, p:24}. We then investigate the effect of varying the scaling factor $\alpha$ in the adaptive patch partitioning scheme, followed by an evaluation of RMS pooling and how different prediction strategies influence overall classification performance.

\begin{figure}[t]
	\centering
	\includegraphics[width=0.9\columnwidth]{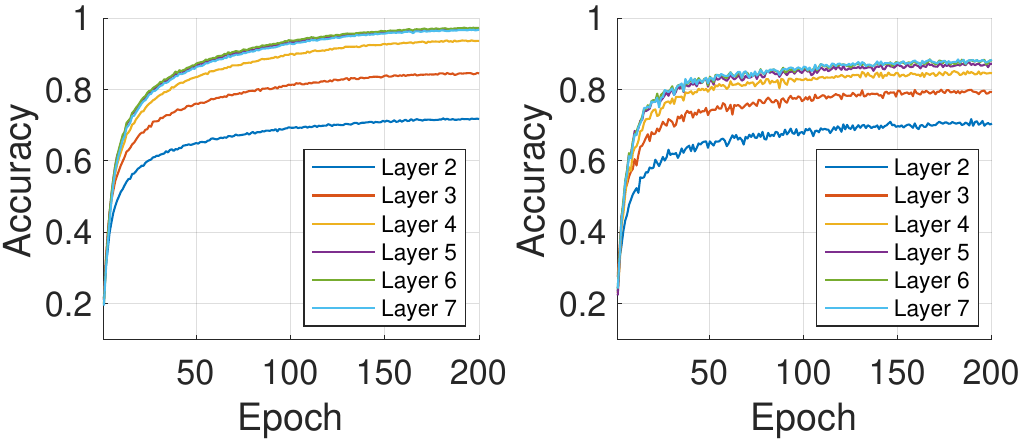} % Reduce the figure size so that it is slightly narrower than the column.
	\caption{Layer-wise prediction accuracy (layers 2-7) of VGG8 trained with ASGE on CIFAR-10. First and final layers excluded. Left: training. Right: validation.}
	\label{fig3}
\end{figure}

\begin{table*}[t]
	\centering
	\begin{tabular}{lccccccc}
		\toprule
		\textbf{ASGE ($\alpha$)} & \textbf{Model} &\textbf{Partition ($P_l$)} &\textbf{MNIST} & \textbf{F-MNIST} & \textbf{CIFAR-10} & \textbf{CIFAR-100}\\
		\midrule
		0.0 & VGG8 & \{1, 1, 1\} & 99.61 $\pm$ 0.01 & 92.88 $\pm$ 0.08 & 88.27 $\pm$ 0.20 & 61.69 $\pm$ 0.23\\
		0.5 & VGG8 & \{2, 1, 1\}& 99.64 $\pm$ 0.03 & 92.79 $\pm$ 0.10 & 89.20 $\pm$ 0.07 & 62.31 $\pm$ 0.29  \\
		\textbf{1.0} & VGG8 & \{\textbf{4, 2, 1}\} & \textbf{99.65} $\pm$ \textbf{0.03} & \textbf{93.41} $\pm$ \textbf{0.14} & \textbf{90.50} $\pm$ \textbf{0.08} & \textbf{65.34} $\pm$ \textbf{0.31} \\
		1.5 & VGG8 & \{6, 3, 1\}& 99.63 $\pm$ 0.01 & 93.36 $\pm$ 0.07 & 90.22 $\pm$ 0.13 & 64.84 $\pm$ 0.35 \\
		2.0 & VGG8 & \{8, 4, 2\}& 99.57 $\pm$ 0.04 & 93.12 $\pm$ 0.10 & 86.91 $\pm$ 0.22  & 60.40 $\pm$ 0.23\\
		\bottomrule
	\end{tabular}
	\caption{Test accuracy (\%) of ASGE under different values of $\alpha$ across four datasets. VGG8 contains layers with $C_l \in $ \{128, 256, 512\} and \emph{Partition} row shows the corresponding $P_l$ assigned to each $C_l$ according to formula (\ref{formula3}).}
	\label{tab1}
\end{table*}

\begin{figure*}[t]
	\centering
	\includegraphics[width=0.8\textwidth]{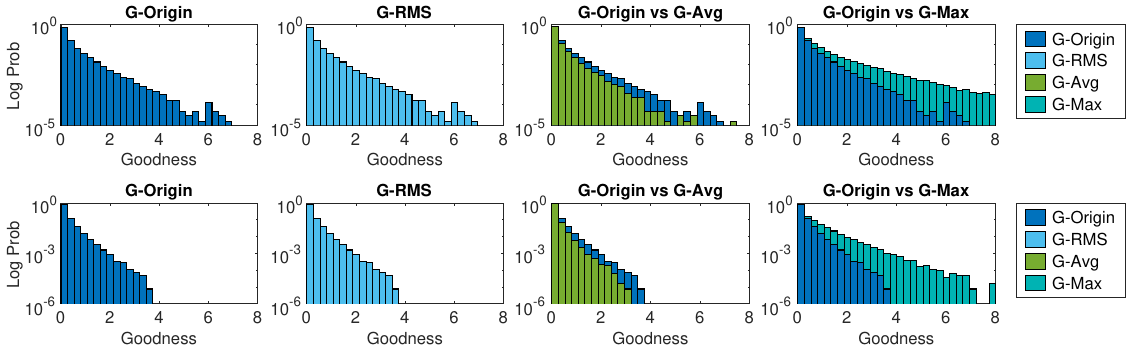} % Reduce the figure size so that it is slightly narrower than the column.
	\caption{Spatial goodness distributions on a logarithmic scale before (G-Origin) and after different pooling (G-RMS, G-Avg, G-Max) in a layer that has pooling. Each row represents a different channel configuration: C = 512 (top) and C = 256 (bottom).}
	\label{fig4}
\end{figure*}

\begin{table*}[t]
	\centering
	\begin{tabular}{lcccccc}
		\toprule
		\textbf{ASGE (Pooling)} & \textbf{Model} &\textbf{MNIST} & \textbf{F-MNIST} & \textbf{CIFAR-10} & \textbf{CIFAR-100} \\
		\midrule
		Average & VGG8 & 99.62 $\pm$ 0.03 & 93.18 $\pm$ 0.10 & 90.35 $\pm$ 0.15 & 64.96 $\pm$ 0.16\\
		Max  & VGG8 & 99.61 $\pm$ 0.03 & 93.07 $\pm$ 0.16 & 90.46 $\pm$ 0.18 & 64.41 $\pm$ 0.11\\
		\textbf{RMS} & \textbf{VGG8} &\textbf{99.65} $\pm$ \textbf{0.03} & \textbf{93.41} $\pm$ \textbf{0.14} & \textbf{90.50} $\pm$ \textbf{0.08} & \textbf{65.34} $\pm$ \textbf{0.31} \\
		\bottomrule
	\end{tabular}
	\caption{Test accuracy (\%) of ASGE with different pooling operations. }
	\label{tab2}
\end{table*}

\begin{table*}[t]
	\centering
	\begin{tabular}{lccrrr}
		\toprule
        \shortstack{\textbf{Prediction} \\ \textbf{Strategy}} &
        \textbf{Model}   & 
        \shortstack{\textbf{Test} \\ \textbf{Accuracy (\%)}}   &
        \shortstack{\textbf{Activation} \\  \textbf{Memory (MB)}} & \shortstack{\textbf{Classifier} \\ \textbf{Params}  ($\times$10\textsuperscript{4})} & \textbf{Best Layer} \\
        \midrule
        BP (repr.) & VGG8 & \textbf{69.11} $\pm$ \textbf{0.31} & 150  & 5.13 & - \\
        ASGE Fusion Pred & VGG8 & 65.34 $\pm$ 0.31 & 1.25 & 25.66 & - \\
        ASGE Last Pred   & VGG8 & 63.31 $\pm$ 0.20 & 0.25 & 5.13 & - \\
        ASGE Best Pred   & VGG8 & 63.60 $\pm$ 0.44 & \textbf{0}    & \textbf{0}                                  & 7 \\
        \bottomrule
	\end{tabular}
	\caption{CIFAR-100: ASGE prediction strategies vs. BP. \emph{Activation Memory}: calculated using a batch of 128. \emph{Best Layer}:  the layer with the highest validation accuracy.}
	\label{tab3}
\end{table*}

\begin{table*}[!t]
    \centering
    \begin{tabular}{llcccccc}
    \toprule
    &  & &  &  & & \multicolumn{2}{c}{\textbf{ImageNet}}\\
    \textbf{Method} & \textbf{Model}& \textbf{MNIST}& \textbf{F-MNIST} & \textbf{CIFAR-10}    &\textbf{CIFAR-100} & \textbf{Top-1} & \textbf{Top-5}\\
    \midrule
    BP (repr.) & VGG8 &  99.64 $\pm$ 0.03 & 94.29 $\pm$ 0.10& 91.68 $\pm$ 0.08 & 69.11$\pm$ 0.31 & 46.16 & 72.02\\
    \midrule
    FA  & MLP8 & 98.74 & - & 58.03 & - & 6.92 & 17.46 \\
    DTP & VGG6 & 98.93 $\pm$ 0.04 & \textbf{90.91} $\pm$ \textbf{0.17} & 85.33 $\pm$ 0.32 &- & 1.66 & 5.44 \\
    DFA & CNN5 & 98.98 $\pm$ 0.05 & - & 73.10 $\pm$ 0.50 & 41.00 $\pm$ 0.30 &6.20&-\\
    %DRTP & CNN3 & 98.52 $\pm$ 0.15 &-& 68.96 $\pm$ 0.45 &-&-&-\\
    SoftHebb & SoftHebb & \textbf{99.35} $\pm$ \textbf{0.03} &-& 80.31 $\pm$ 0.14 & \textbf{56.00} &27.30&-\\
    rLRA  & ResNet18 & 98.18 & 88.13 & \textbf{93.58} &-& \textbf{69.22} & \textbf{87.96}\\
    PEPITA &CNN2 & 98.29 $\pm$ 0.13 & - & 56.33 $\pm$ 1.35 & 27.56 $\pm$ 0.60 &-&-\\
    %LLS\textsubscript{square} & VGG8 & \textbf{99.54} $\pm$ \textbf{0.01} & \textbf{93.54} $\pm$ \textbf{0.06} & 88.64 $\pm$ 0.12 & \textbf{58.84} $\pm$ \textbf{0.33} &-&- \\
    \midrule
     FF & MLP4 & 98.69 &-&59.00&-&-&- \\
     CaFo & CNN3 & 98.80 & - & 67.43 & 40.76 &-&-\\  
     SymBa & MLP3 & 98.58 & - & 59.09 & 29.28 &-&-\\
     CwC & CNN4 & 99.42 $\pm$ 0.08 & 92.31 $\pm$ 0.32 & 78.11$\pm$ 0.44 & 51.23 &-&-\\
     DeeperForward & ResNet18 & 99.63 $\pm$ 0.04 & 93.13 $\pm$ 0.13 & 88.72 $\pm$ 0.17 & 53.09 $\pm$ 0.79 &-&-\\
     \textbf{ASGE (ours)} & VGG8 & \textbf{99.65} $\pm$ \textbf{0.03} & \textbf{93.41} $\pm$ \textbf{0.14} & 90.50 $\pm$ 0.08 & 65.34 $\pm$ 0.31 & 23.04 & 43.08\\
     \textbf{ASGE (ours)} & VGG11 & 99.59 $\pm$ 0.06 & 92.98 $\pm$ 0.24 & \textbf{90.62} $\pm$ \textbf{0.17} & \textbf{65.42} $\pm$ \textbf{0.16} & 26.21 & 47.49\\
     \textbf{ASGE(ours)} & ResNet18-CHx4 & - & - & - & - & \textbf{51.58} & \textbf{75.23} \\
    \midrule
    
    \end{tabular}
    \caption{Test accuracy (\%) comparison: ASGE vs. standard BP, non-FF and FF-based BP-free methods. ResNet18-CH×4 uses 4$\times$ wider channels than standard ResNet-18.}
    \label{tab4}
\end{table*}

\subsubsection{Effect of ASGE}

Figure~\ref{fig1} demonstrated a core limitation of FF training: a progressive degradation of layer-wise prediction accuracy with increasing network depth in CNNs. To assess whether ASGE mitigates this issue, Figure~\ref{fig3} presents the layer-wise accuracy of our VGG8 architecture, excluding the first layer---due to its negligible predictive performance---and the final linear classifier. The results reveal a fundamental reversal: from layers 2--7, prediction accuracy consistently improves across both the training and validation sets. Notably, the inter-layer accuracy gain diminishes with depth, indicating stabilization of increasingly refined feature representations. Overall, these findings confirm that ASGE successfully counteracts the FF's representational degradation, enabling effective information accumulation across layers for scalable deep network learning.

\subsubsection{Effect of Partitioning Factor $\alpha$}

To examine the impact of the hyperparameter $\alpha$ on ASGE performance, we evaluate five representative values: 0, 0.5, 1, 1.5, and 2. All other training configurations are held constant, including the use of RMS pooling and the \emph{Fusion Pred} strategy for final prediction. The results are summarized in Table~\ref{tab1}. Across all datasets, a consistent trend emerges: test accuracy improves as $\alpha$ increases from 0 to 1, but begins to decline when $\alpha$ exceeds 1. The best performance is achieved at $\alpha$ = 1, yielding test accuracies of 99.65\% on MNIST, 93.41\% on F-MNIST, 90.50\% on CIFAR-10, and 65.34\% on CIFAR-100. We interpret this behavior as follows: increasing $\alpha$ results in more spatial partitions per layer, allowing each layer to capture richer and more fine-grained spatial goodness information, which enhances performance. However, when $\alpha$ grows too large, the surplus patches introduces redundant information and noise, causing potential overfitting and degraded generalization. Thus, $\alpha$ = 1 provides an effective balance between spatial granularity and noise, enabling optimal performance.

\subsubsection{Effect of RMS Pooling}

Figure~\ref{fig4} visualizes the impact of different pooling operations on spatial goodness distributions. The first two columns show the distributions from the original representation (G-Origin) and after applying RMS pooling (G-RMS). The subsequent two columns compare G-Origin with the distributions obtained after average pooling (G-Avg) and max pooling (G-Max), respectively. It is evident that G-RMS closely mirrors G-Origin, preserving both the tail behavior and the overall distribution structure. In contrast, G-Avg distorts the original distribution by suppressing the tail and shifting toward lower goodness scores. G-Max,  on the other hand, exaggerates the tail, increasing the proportion of higher goodness values. This indicates that RMS pooling more faithfully preserves spatial goodness information during downsampling, whereas the other two pooling methods introduce distributional distortions. Furthermore, we train a VGG8 using each pooling method with $\alpha$ = 1 and the \emph{Fusion Pred} strategy for final prediction. As shown in Table~\ref{tab2}, ASGE with RMS pooling consistently outperforms average and max pooling across all datasets. By more efficiently preserving goodness information, RMS pooling aligns better with the core principles of ASGE and enhances the model's representational capacity.

\subsubsection{Performance of Prediction Strategies}We evaluate the three proposed strategies for final prediction against standard BP using the same VGG8 with $\alpha$ = 1 and RMS pooling, measuring test accuracy, classifier parameters, and intermediate activation memory. The only architectural modification occurs in \emph{Best Pred}, which does not require a final classifier. Thus, its last layer is replaced with a copy of the previous convolutional layer. As shown in Table~\ref{tab3}, \emph{Fusion Pred} achieves the highest accuracy among the three (65.24\%), coming closest to that of BP, but at the cost of the largest number of classifier parameters and the highest intermediate activation memory, though still requiring 120$\times$ less memory than BP. In contrast, \emph{Best Pred} sacrifices only 1.74\% accuracy compared to \emph{Fusion Pred}, while completely eliminating the final classifier and storing intermediate activations. Since the seventh layer yields the best prediction, inference can terminate after this layer, further reducing computational overhead. \emph{Last Pred} uses the same number of classifier parameters as BP and consumes 600$\times$ less activation memory, but delivers the lowest accuracy. \emph{Last Pred} underperforms due to less discriminative raw activations under ASGE's goodness supervision. In summary, the proposed strategies offer a flexible trade-off between accuracy, memory consumption, and computational cost, enabling efficient deployment across various resource-constrained scenarios.

\subsection{Comparison with Other Work}We benchmark ASGE against standard BP and diverse BP-free approaches, including both FF-based and non-FF methods. Among non-FF methods, we include FA \cite{s:18}, Difference Target Propagation (DTP) \cite{ma:22, s:18}, DFA \cite{a:16,b:19}, SoftHebb \cite{a:23}, rLRA \cite{o:23} and PEPITA \cite{g:23}). For FF-based methods, we evaluate the original FF \cite{g:22}, CaFo \cite{z:23}, SymBa \cite{h:23}, CwC \cite{p:24} and DeeperForward \cite{l:25}. All experiments for ASGE are conducted using VGG8 and VGG11 with $\alpha$ = 1, RMS pooling, and \emph{Fusion Pred} for final prediction.

Table~\ref{tab4} shows that ASGE consistently outperforms prior FF-based methods across MNIST, F-MNIST, CIFAR-10, and CIFAR-100, and surpasses most non-FF BP-free methods. The only exception is rLRA, which achieves a 2.96\% higher accuracy on CIFAR-10. Notably, on the more challenging CIFAR-100 dataset, ASGE shows a significant advantage, exceeding all previous FF-based methods by at least 12.33\%. Compared to BP, ASGE delivers slightly lower but still competitive results, with accuracy gaps of just 0.88\% on F-MNIST, 1.18\% on CIFAR-10, and 3.77\% on CIFAR-100. 

To our knowledge, ASGE is the first FF-based method to successfully scale to ImageNet, achieving Top-1 and Top-5 accuracies of 26.21\% and 47.49\%, setting a new benchmark for FF-based training. ASGE outperforms non-FF BP-free methods like FA, DTP, and DFA, though it trails SoftHebb, rLRA, and standard BP. However, rLRA requires a full forward pass before updates, increasing memory usage, while SoftHebb is limited to specific network architectures. ASGE circumvents these constraints with fully layer-wise training and flexible integration with convolutional architectures.

\section{Conclusion}
This work introduces ASGE, a novel FF-based BP-free training framework for CNNs that overcomes key limitations of prior FF methods through channel-aware spatial goodness supervision, enhancing both representational capacity and scalability to large-scale datasets. Experiments demonstrate that ASGE consistently outperforms all FF-based methods across 5 image classification benchmarks, and further narrows the performance gap with BP. Notably, it marks the first successful application of FF-based training to ImageNet, demonstrating its potential as a scalable and effective BP-free paradigm for modern CNNs.

\subsubsection{Limitations}Despite its strengths, ASGE has certain limitations. As a greedy, layer-wise approach, it generally requires longer training time to converge than end-to-end BP. Currently, ASGE is only applied to CNNs, and its extension to other architectures such as Transformers or RNNs remains to be explored. Additionally, while ASGE performs competitively across various datasets, it still trails some non-FF BP-free methods on ImageNet, indicating room for future improvement.

\bibliography{aaai2026}

\clearpage

\section{Technical Appendix}
\subsection{Model Architecture} In this work, we use two VGG-like CNNs, which are VGG8 and VGG11. We replace all fully connected layers---except for the final classification layer---with convolutional layers. Each convolution block (ConvBlock) comprises a convolution operation, a ReLU activation function, RMS pooling (if present) and layer normalization. A dropout of 0.1 is applied before layer normalization. The layer normalization is implemented as root mean square (RMS) normalization, which normalizes activations by their RMS. The model configurations of VGG8 and VGG11 are summarized in Table~\ref{tab5}.

\begin{table}[h]
    \centering
    \begin{tabular}{ccc}
        \toprule
        \textbf{Block} & \textbf{VGG8} & \textbf{VGG11}\\
        \midrule
        1 & \shortstack{ConvBlock\\3, 128, 1} & \shortstack{ConvBlock \\3, 128, 1}\\
        \midrule
        2 & \shortstack{ConvBlock (Pool)\\3, 256, 1} & \shortstack{ConvBlock (Pool)\\3, 256, 1} \\
        \midrule
        3 & \shortstack{ConvBlock\\3, 256, 1} & \shortstack{ConvBlock\\3, 256, 1} \\
        \midrule
        4 & \shortstack{ConvBlock (Pool)\\3, 512, 1} & \shortstack{ConvBlock (Pool)\\3, 512, 1} \\
        \midrule
        5 & \shortstack{ConvBlock (Pool)\\3, 512, 1} & \shortstack{ConvBlock \\3, 512, 1} \\
        \midrule
        6 & \shortstack{ConvBlock (Pool)\\3, 512, 1} & \shortstack{ConvBlock (Pool)\\3, 512, 1} \\
        \midrule
        7 & \shortstack{ConvBlock \\3, 512, 1} & \shortstack{ConvBlock \\3, 512, 1} \\
        \midrule
        8 & \shortstack{Linear \\ 10} & \shortstack{ConvBlock (Pool)\\3, 512, 1} \\
        \midrule
        9 & & \shortstack{ConvBlock\\3, 512, 1} \\
        \midrule
        10 &  & \shortstack{ConvBlock\\3, 512, 1} \\
        \midrule
        11 &  & \shortstack{Linear\\10} \\
        \bottomrule
       
    \end{tabular}
    \caption{Model configurations for 10 classes. The ConvBlock entries show kernel size, output channels and stride. }
    \label{tab5}
\end{table}

\subsection{Data Augmentation}
\subsubsection{MNIST}All training, validation and test images are resized to 32$\times$32. 
\subsubsection{F-MNIST}Training images are resized to 32$\times$32, followed by random horizontal flipping. Validation and test images are resized to 32$\times$32.
\subsubsection{CIFAR-10 \& CIFAR-100} We apply a richer set of data augmentations to training images, including random cropping to 32$\times$32 with 4-pixel padding, random horizontal flipping, color jittering (brightness 0.4, contrast 0.4, saturation 0.1, hue 0.1), and random grayscale conversion with a probability of 0.2.
\subsubsection{ImageNet}During training, each image is randomly resized and cropped to a resolution of 224$\times$224, followed by random horizontal flipping, color jittering (brightness: 0.4, contrast: 0.4, saturation: 0.4, hue: 0.1) and normalization.
For both test and validation phases, all images are first resized to  256$\times$256, then center-cropped to 224$\times$224, and finally normalized.

\subsection{Hyperparameters} We present the hyperparameter searches for non-ImageNet (Table~\ref{tab6}) datasets and for ImageNet (Table~\ref{tab7}) due to differences in scale. We choose hyperparameters according to the validation accuracy.
\begin{table}[h]
    \centering
    \begin{tabular}{ccc}
    \toprule
     Hyperparameters & Range & Chosen\\
     \midrule
     Optimizer & SGD, AdamW & AdamW \\
     \midrule
     Scheduler & Cosine, ROP & Cosine\\
     \midrule
     Batch Size & 64, 128 & 128\\
     \midrule
     \shortstack{Initial \\ learning rate}  & \shortstack{[0.0001, 0.001]\\step:0.0001} & 0.0002\\
     \midrule
     \shortstack{Ending \\ learning rate} & \shortstack{[0.00001, 0.00005]\\step:0.00001} & 0.00001\\
     \midrule
     \shortstack{Training Epoch:\\ (F-) MNIST} & 50, 100, 150 & 100\\
     \midrule
     \shortstack{Training Epoch:\\ CIFAR-10 (100)} & 200, 300, 400 & 400\\

     \bottomrule
    \end{tabular}
    \caption{Hyperparameter search for non-ImageNet datasets. ROP: ReduceLROnPlateau}
    \label{tab6}
\end{table}

\begin{table}[h]
    \centering
    \begin{tabular}{lcc}
     \toprule
     Hyperparameters & Range & Chosen\\
     \midrule
     Optimizer & SGD, AdamW & SGD \\
     \midrule
     Scheduler & \shortstack{Cosine\\ ReduceOnPlateau} & Cosine\\
     \midrule
     Batch Size & 256, 512 & 512\\
     \midrule
     \shortstack{Initial \\ learning rate}  & 0.2, 0.1, 0.05, 0.01 & 0.05\\
     \midrule
     \shortstack{Ending \\ learning rate} & 0.0005, 0.0001 & 0.0005\\
     \midrule
     \shortstack{Training \\ Epoch} & 45,90 & \shortstack{45: VGG \\ 90: ResNet}\\
     \bottomrule
    \end{tabular}
    \caption{Hyperparameter search for ImageNet.}
    \label{tab7}
\end{table}

\subsection{Reproduced Results}
For the reproduction of the original Forward-Forward (FF) on VGG8, positive samples are the original images. For negative samples, two randomly selected input images are combined using a random binary mask. Specifically, one sample is multiplied element-wise by a random mask of 0s and 1s, while the other is multiplied by the complement (1 minus the mask). The two masked images are then added to generate a synthetic negative sample. The threshold for determining whether a sample is positive or negative is set to the total number of activations in the feature maps ($C_l \times H_l \times W_l$).

For the reproduction of backpropagation (BP), each ConvBlock is composed of a convolution operation, a batch normalization, a ReLU activation function and average pooling (if present).

\end{document}